\documentclass[conference,dvipsnames]{IEEEtran}
\IEEEoverridecommandlockouts
\usepackage{cite}
\usepackage{amsmath,amssymb,amsfonts}
\usepackage{algpseudocode}
\usepackage{graphicx}
\ifCLASSOPTIONcompsoc
  \usepackage[caption=false,font=normalsize,labelfont=sf,textfont=sf]{subfig}
\else
  \usepackage[caption=false,font=footnotesize]{subfig}
\fi
\usepackage{textcomp}
\usepackage{xcolor}
\usepackage{todonotes}
\usepackage{booktabs}
\usepackage{url}
\usepackage{tikz}
\usetikzlibrary{trees, matrix, positioning, patterns, shapes, shadows, shapes.arrows, arrows.meta, shapes.multipart, decorations.pathreplacing, calc, tikzmark, shapes.geometric}
\usepackage{pgfplots}
\pgfplotsset{compat=1.17}
\usepackage{pgf-pie}
\usepackage[binary-units]{siunitx}
\usepackage{mathdots}
\usepackage{mathtools}
\usepackage{bm}
\usepackage[colorlinks=true, linkcolor=black, urlcolor=NavyBlue, citecolor=black]{hyperref}

\def\BibTeX{{\rm B\kern-.05em{\sc i\kern-.025em b}\kern-.08em
    T\kern-.1667em\lower.7ex\hbox{E}\kern-.125emX}}
\begin{document}

\title{Compressing the Backward Pass of Large-Scale Neural Architectures by Structured Activation Pruning}

\author{
\IEEEauthorblockN{Daniel Barley and Holger Fröning}
\IEEEauthorblockA{\textit{Computing Systems Group, Institute of Computer Engineering} \\
\textit{Heidelberg University, Germany}\\
daniel.barley@ziti.uni-heidelberg.de,  
holger.froening@ziti.uni-heidelberg.de}
}

\maketitle

\begin{abstract}
  The rise of Deep Neural Networks (DNNs) has led to an increase in model size
  and complexity, straining the memory capacity of GPUs. Sparsity in DNNs,
  characterized as structural or ephemeral, has gained attention as a solution.
  This work focuses on ephemeral sparsity, aiming to reduce memory consumption
  during training. It emphasizes the significance of activations, an often
  overlooked component, and their role in memory usage. This work employs
  structured pruning in Block Sparse Compressed Row (BSR) format in combination
  with a magnitude-based criterion to efficiently prune activations. We
  furthermore introduce efficient block-sparse operators for GPUs and showcase
  their effectiveness, as well as the superior compression offered by block
  sparsity. We report the effectiveness of activation pruning by evaluating
  training speed, accuracy, and memory usage of large-scale neural
  architectures on the example of ResMLP on image classification tasks. As a
  result, we observe a memory reduction of up to 32\% while maintaining
  accuracy. Ultimately, our approach aims to democratize large-scale model
  training, reduce GPU requirements, and address ecological concerns.
\end{abstract}

\section{Introduction}\label{sec:introduction}

Deep Neural Networks (DNNs) have grown exponentially in size and complexity in
recent years. Even though the success of deep learning is, in part, fueled by
advancements in Graphics Processor Unit (GPU) technology, this rapid increase
in model size is in direct conflict with the notoriously small memory capacity
of GPU architectures. Resource efficiency in
general~\cite{10.1145/3587095,roth2020} and
sparsity~\cite{DBLP:journals/corr/abs-1810-05270} in particular has therefore
become an active field of research in deep learning. There are many approaches
to sparsity in DNNs. Fundamentally they can be categorized into
\emph{structural}, or \emph{model} sparsity, and \emph{ephemeral} sparsity.
Structural sparsity is applied to parts of the network itself, this may range
from individual weights~\cite{precheltConnectionPruningStatic1997} to whole
layers~\cite{schindlerParameterizedStructuredPruning2019}. The sparse pattern
is then inherent to the network and applied equally to all inputs. Structurally
sparse approaches, like magnitude-based weight pruning, require information on
network parameters unknown prior to training. Ephemeral sparsity, on the other
hand, is applied dynamically per input. This allows for pruning criteria
independent of network parameters. Because network parameters are subject to
constant change during the training phase of a DNN, most structural approaches,
aside from simple static masks, disqualify for training. An ephemeral pruning
strategy is therefore pursued in this work.

With the type of sparsity in place, the sparse component remains to be
determined. One can observe that the state of neural architectures is composed
of weights, biases, gradients, and activations. Activations are often overseen
as they seem to exist only temporarily during forward inference. During
training, however, the backpropagation algorithm requires them to be stored
until the backward path is executed. Moreover, activations make up the vast
majority of the memory footprint during training as they scale with an
additional mini-batch dimension. For the targeted ResMLP
architecture~\cite{touvronResMLPFeedforwardNetworks2021} we observe that 90\%
of the total memory consumption is due to activations. This percentage is even
larger for convolutional architectures, as weight sharing reduces memory
pressure with regard to parameters, effectively increasing the percentual
contribution made by activations to the overall state.
Considering Amdahl's Law, we identify activations to offer the largest
potential for reducing state.

\begin{table}[!t]
  \centering
  \caption{ResMLP-S12 Memory Footprint by Component}\label{tab:component}
  \begin{tabular}{lcccc|c}
    \toprule
    Component & Input & Model & Optimizer & Activations & Total\\
    \midrule
    Memory [\si{\mebi\byte}] & 18.5 & 59.6 & 121.1 & 1599.2 & 1798.4 \\
    \% & 1.0 & 3.3 & 6.7 & 89.0 & 100.0 \\
    \bottomrule
  \end{tabular}
\end{table}

While pruning has become a widely used method, it is equally well known that
most processor architectures do not handle unstructured sparsity well. This is
especially true for GPUs, where irregular memory access patterns lead to a
rapid degradation of memory bandwidth. Thus, the present work considers pruning
in a structured manner, so that the number of activations required to be stored
is reduced substantially, while the resulting tensor shapes are inline with an
efficient execution on GPUs. The Block Sparse Compressed Row (BSR) format we
opted for in this work has the additional benefit of a better compression
ratio for increasing block sizes.

Furthermore, we use a simple magnitude-based criterion to select activations during the
forward pass and prune them only after the computation is completed. This
ensures an accurate loss computation and avoids problems like vanishing
activations. The $l2$-norm is used to compare blocks of activations, which
are then ranked using an efficient top-k approach, pruning the lowest
scoring blocks.

One of the challenges faced exploiting structured sparse activations for
training is the lack of suitable block-sparse operators in publicly available
libraries. We therefore present a set of suitable operators with support for
BSR-compressed activations. Using these, we can assess the effectiveness of
block-sparse activation pruning in terms of model accuracy, training time, and
reduction of state, the latter being the main focus of this work.

In summary, this work makes the following contributions:
\begin{enumerate}
  \item We introduce a set of efficient block-sparse operators for GPU
    architectures, which combine comparable performance with dense linear
    algebra libraries but support as low as 30\% of sparsity efficiently.
  \item We propose a simple yet effective method to prune activations in a
    structured manner, creating sparsity patterns inline with the previous
    operators.
  \item We evaluate training speed, accuracy, and memory footprint of
    large-scale neural architectures for image classification in comparison to
    baseline implementations, highlighting the effectiveness of the activation
    pruning concept.
\end{enumerate}

As we will see in the following, the magnitude-based activation pruning method
is simple yet surprisingly effective. Overall, we see this work as a first step
in democratizing the training of large-scale models in a way that it becomes
more accessible for researchers without access to large-scale training
clusters. Furthermore, requiring less GPUs for a given training task is
reducing the associated carbon footprint, which is of increasing importance
given the trend towards larger and larger models while ecological implications
are becoming more severe.

\section{Background and Related Work}

\subsection{Sparse data structures}

Sparse linear algebra has a long tradition in many disciplines like scientific
computing, data science, and, of course more recently, machine learning. As a
result, there exists a plethora of sparse matrix formats and libraries,
tailored to the specific needs of the application. Compressing sparse matrices
is achieved by storing the non-zero elements only, alongside metadata to
reconstruct the original matrix. Well known formats like \texttt{Compressed Sparse
Row/Column (CSR/CSC)} formats use lists of corresponding row and column indices of
non-zero values to that end. To be viable, the ratio of zero
elements to the total number of values, or sparsity, has to be sufficiently
large. If sparsity is very low, the additional encoding overhead may even cause
memory consumption to be larger than saving the original matrix. For CSR,
which is natively supported by PyTorch, sparsity has to be larger than 50\%
to effectively achieve compression, otherwise the coding overhead will dominate. 
As will be discussed later in this work, for sparse
training we can only expect sparsity up to around 80\%.

Another format supported by PyTorch is the \texttt{Block Sparse Compressed Row (BSR)}
format. BSR works analogously to CSR, the difference being that BSR does not
encode individual values, but rather blocks of values. Like with CSR, an array
of compressed row indices (\texttt{crow}) is used to encode the cummulative
number of non-zero blocks per row of the original matrix. The last element in
\texttt{crow} is then the total number of non-zero blocks. To get the number of
non-zero blocks for a specific row $n$ we subtract the $n$-th \texttt{crow}
entry from the $n+1$-th. To then get the block's position within a row the
column index (\texttt{col}) array is accessed at the value of
$\texttt{crow}[n]$. In case of multiple blocks per row, following \texttt{col}
entries encode their positions. The larger the BSR block size, the fewer
indices are required, improving compression. Table~\ref{tab:bsr_compression}
shows the additional encoding overhead of BSR for various block sizes $b$ at
increasing sparsity $s$. In addition to better compression, block-sparse formats
complement the warp execution paradigm of GPUs very well. For sufficiently
large blocks load and store operations can be coalesced and a high throughput
thereby maintained.

While PyTorch offers a BSR tensor construct, most operators do not yet
support it, hence there is a need for custom compute kernels.

\begin{table}[!t]
	\centering
	\caption{BSR Memory Overhead Over Ideal Compression \(1 -
	s\)}\label{tab:bsr_compression}
	\begin{tabular}{lrrrrrrrr}
	\toprule
	\(b\) & 1 & 4 & 8 & 16 & 32 & 64 & 128 & 384 \\
    $s$ [\%] & \multicolumn{7}{c}{Additional Encoding Overhead [\%]}  \\
	\midrule
	0 & 100.26 & 25.26 & 12.76 & 6.51 & 3.39 & 1.82 & 1.04 & 0.52 \\
	20 & 80.26 & 20.26 & 10.26 & 5.26 & 2.78 & 1.53 & 0.82 & 0.57 \\
	40 & 60.26 & 15.26 & 7.76 & 4.00 & 2.13 & 1.23 & 0.76 & 0.62 \\
	60 & 40.26 & 10.26 & 5.26 & 2.77 & 1.52 & 0.85 & 0.54 & 0.16 \\
	80 & 20.26 & 5.26 & 2.77 & 1.52 & 0.87 & 0.56 & 0.49 & 0.21 \\
	100 & 0.26 & 0.26 & 0.26 & 0.26 & 0.26 & 0.26 & 0.26 & 0.26 \\
	\bottomrule
	\end{tabular}
\end{table}

\subsection{Neural architectures for computer vision}

The Vision Transformer (ViT)~\cite{dosovitskiyImageWorth16x162021} is a
state-of-the-art deep learning model that uses the Transformer architecture to
analyze images by breaking them into patches and processing them as sequences.
Its self-attention mechanism makes it very effective for tasks like image
classification. However, assessing the effectiveness of activation pruning on
ViT would require substantial engineering effort due to various operators found
in this neural architecture, and, as previously reported, block-sparse
operators are not supported natively by standard tool stacks.

Thus, the ResMLP architecture proposed by Touvron \textit{et
al.}~\cite{touvronResMLPFeedforwardNetworks2021} was chosen for
experimentation for two main reasons: (1) It is a rather simple architecture
consisting of linear layers only, which lowers the engineering overhead for
custom operators. (2) As a descendant of the Vision Transformer, it employs a
similar embedding scheme, processing patches of the input image which naturally
defines a hardware friendly structure for pruning.

\begin{figure}[!t]
  \centering
  \includegraphics[]{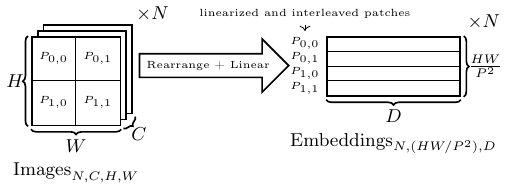}
	\caption{Patch embeddings used in ResMLP and other vision MLPs. The input
	image ($(N, C, H, W)$ tensor) is divided into patches $P_{i,j}$ which are then
	linearized and color-channel interleaved. Linearized patches are finally
	scaled to a configurable hidden dimension $D$ via a linear
	layer.}\label{fig:embeddings}
\end{figure}

The input image is split into rectangular patches which are then linearized. In
case of color images, color channels are interleaved. The resulting row vector
is then scaled to a hidden dimension consistent throughout the network. For the
full image we obtain an embedding matrix with rows corresponding to patches,
as shown in Fig.~\ref{fig:embeddings}. From this embedding matrix we can easily prune
rows, or at least parts thereof. Because rows are contiguous in memory, no
rearrangement of data is needed to encode the pruned embedding matrix to BSR.
The underlying intuition is that we only want to group values that have a
physical relationship.

\subsection{Related work}

Reviewing other ephemeral approaches to sparsity helped with the development
tremendously. One of the most well known examples is
\emph{dropout}~\cite{hintonImprovingNeuralNetworks2012}. Originally developed
as a regularizer, dropout randomly forces activations to zero, sparsifying the
connections in a layer. This aims to prevent co-adaption, where feature
detectors strongly correlate with other feature detectors, making them useless
on their own, which also prevents overfitting. 
In its original form, dropout
induces unstructured sparsity, which is, as established, not suited for
compression. There exists a
DropBlock~\cite{ghiasiDropBlockRegularizationMethod2018} variant, which drops
spatially contiguous blocks. However, leaving the forward pass unaltered was
one of the goals of this work.

Many alternative approaches seek to prune gradients. Sparse gradients are
especially beneficial in the context of distributed training, where they have
to be broadcasted among compute ranks. Most approaches also prune based on
magnitude, using either fixed thresholds or adaptive ones obtained via
reduction. Zhang \textit{et al.} propose a technique called Memorized Sparse
Backpropagation~\cite{zhangMemorizedSparseBackpropagation2020a}. The network
essentially estimates a gradient that is empirically shown to be sufficiently
similar to the real gradient to effectively train. Unpropagated gradients are
still stored for future updates to counteract information loss. While this
reduces communication, memory is not saved in this approach. We
largely adapt the top-k pruning strategy from this work, however.

A more sophisticated pruning criterion is presented by Liu \textit{et
al.}~\cite{liuDynamicSparseTraining2020}. Their Dynamic Sparse Training (DST)
strategy turns pruning thresholds into learnable parameters. Because the model
has to experiment with different configurations, DST uses soft pruning, meaning
pruned values are not actually discarded, but masked instead for future use.
Combined with the fact that DST focuses on weights this makes this strategy
inapplicable to our work.

The closest to our work is Sparse Weight Activation Training
(SWAT)~\cite{raihanSparseWeightActivation2020}. SWAT uses a very similar
approach for pruning activations in the context of CNNs, but not with the goal
of compression, but faster execution on sparse processor architectures instead.
In addition to activation pruning SWAT also adaptively masks weights during the
forward computation, similar to dropout. While structured pruning is mentioned,
the effects of different granularities of said structure are not explored. As
mentioned SWAT focuses on special sparse accelerators in the context of latency
based on hardware simulations, while we focus on memory compression and
efficient execution on GPUs.

\section{Block Sparse Operators}\label{sec:operators}

In this section we discuss the design and implementation of the block-sparse
linear operator developed in the context of this work. While a linear operator
is presented here, the technique used can be extended to any operator in
principle. Additionally, we report performance for possible configurations
using input shapes characteristic to vision MLPs. The benchmarks are performed
on an Nvidia A30 GPU\footnote{3574 CUDA cores at \SI{10.3}{TF\per\second} peak
FP32, \SI{24}{\giga\byte} HBM2 at \SI{933}{\giga\byte\per\second},
\SI{165}{\watt} TDP, PCIe 4.0 x16,
\url{https://www.nvidia.com/en-us/data-center/products/a30-gpu}}.

\begin{figure}[!t]
  \centering
  \includegraphics[]{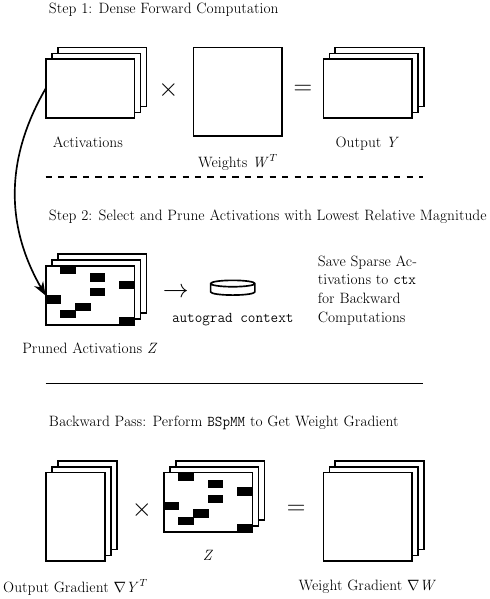}
  \caption{High-level overview of the block-sparse linear operator. After the
  dense forward computation is completed, blocks of activations are pruned based
  on magnitude (top-k pruning). The pruned activations are then saved for the
  backward pass. During the backward pass they are used to calculate the weight
  gradient via block-sparse matrix multiplication (BSpMM)}\label{fig:operator}
\end{figure}

The block-sparse linear operator is designed to be a plug-in replacement for
PyTorch's
\texttt{nn.linear}\footnote{\url{https://pytorch.org/docs/stable/generated/torch.nn.Linear.html}}
layer with two additional parameters: the level of sparsity, and the block size
of the sparse structure. As mentioned in section~\ref{sec:introduction} the
forward computation is kept dense to ensure an accurate loss computation. After
completion of the forward computation, the smallest blocks of input activations
are selected using a magnitude based criterion described in further detail in
section~\ref{sec:pruning}. The selected blocks are forced to zero and the
activation tensor is converted to the BSR format. Fig.~\ref{fig:operator} shows
an overview.

\begin{figure}[!t]
  \centering
  \subfloat[Load BSR indices from global memory.]
  {\includegraphics{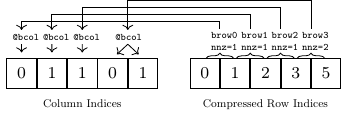}}\label{fig:bsr_groups_a}
  \hfil
  \subfloat[On matching index load non-zero values to registers.]
  {\includegraphics{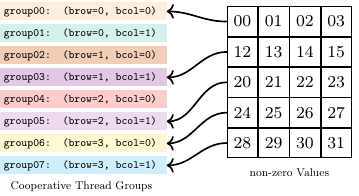}}\label{fig:bsr_groups_b}
  \hfil
  \subfloat[Cache in shared memory]{\includegraphics{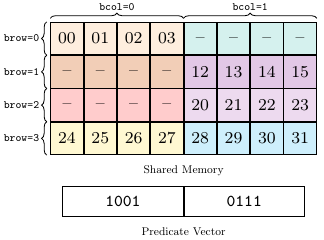}}
  \caption{Cooperative thread groups loading a $4 \times 8$ BSR encoded tensor
  with block size $b=4$ to local shared memory. (a): First parse compressed row
  indices to get the number of non-zero blocks (\texttt{nnz}) in the respective
  row. Use \texttt{nnz} to access the column index array. (b): If row and
  column indices found in (a) match the group load corresponding block from
  global memory to registers. (c): Store loaded values to shared memory for
  following dot product calculation. We use a bit vector to mark zero blocks in
  shared memory, so they are skipped during the compute phase Because the
  number of non-zero blocks is, a priori, unknown enough shared memory has to
  be allocated to fit
  a fully dense block in the worst case.}\label{fig:bsr_groups}
\end{figure}

A custom CUDA block-sparse matrix multiply (BSpMM) kernel is used to calculate
the weight gradient during the backward pass. Because the input and bias
gradients do not depend on activations, they can be computed normally using
native PyTorch operations. The BSpMM implementation is heavily inspired by
CUTLASS\footnote{\url{https://github.com/nvidia/cutlass}}. The kernel can be
divided into a load and a compute phase. We use vectorized loads and
transposition of fragments in shared memory during the load phase to ensure
high throughput. To parse the activation's sparse structure, threads are
grouped into cooperative groups with each group being responsible for a BSR
block. Each group checks if their respective block is non-zero. If it is zero,
the group will skip the load and mark a shared memory internal predicate bit.
During the compute phase threads will skip the corresponding dot product
calculation, if the predicate vector indicates a zero block. See
Fig.~\ref{fig:bsr_groups} for a visualization of the loading process.

\begin{figure}[!t]
  \centering
  \includegraphics[]{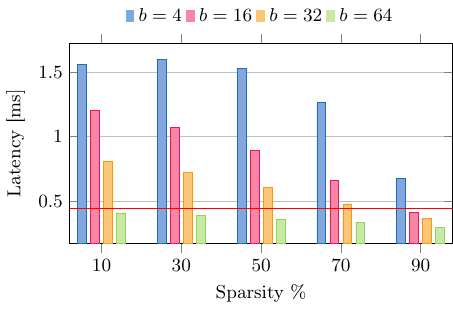}
  \caption{Latency of the BSpMM operator for inputs of shape \((64, 196,
  384)\), different block sizes \(b\), and increasing levels of sparsity. The
  red line marks the latency of PyTorch's dense \texttt{torch.matmul} routine.
  }\label{fig:latency_blocksize}
\end{figure}

The BSR block size has large implications on the performance of the kernel.
Smaller blocks lead to more indices loaded from global memory, lowering the
kernel's FLOP per byte ratio. Also, due to the warp execution style of the
GPU, thread divergence may occur at group sizes smaller than the canonical warp
size of 32. Another, less obvious, limitation occurs at large BSR block sizes.
Because of the limit of 1024 threads per CUDA thread block it becomes necessary
to split the logical cooperative groups among multiple thread blocks. Because
communication among thread blocks is not allowed, indices for the same logical
group will have to be fetched by every participating thread block individually
from global memory, negating the benefit of fewer total indices. Generally,
increasing sparsity decreases latency. However, performance gains
diminish for larger block sizes. At block size 4, the overhead due to
decoding makes up a large percentage of the overall execution time. Therefore,
lowering this overhead via increased sparsity has a greater effect and vice
versa for larger block sizes.  
Fig.~\ref{fig:latency_blocksize} shows the operator's latency at
different block sizes for increasing levels of sparsity.

\begin{figure}[!t]
  \centering
  \includegraphics{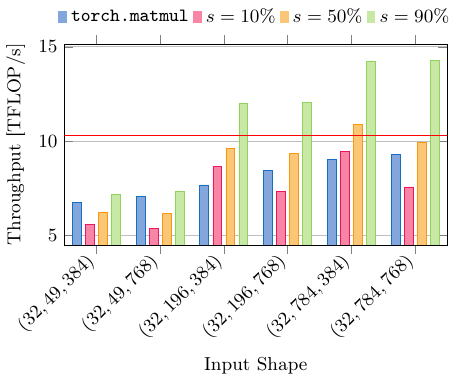}
  \caption{BSpMM throughput at block size 64 compared to dense
  \texttt{torch.matmul} for common input shapes encountered in ResMLP. Note
  that FLOPs refers to the dense number of operations in this context.
  The peak FP32 performance of the A30 GPU used is shown in red.
  }\label{fig:throughput}
\end{figure}

Measuring throughput at block size 64 for input shapes corresponding to ImageNet,
as shown in Figure~\ref{fig:throughput}, shows slight speedups over the dense
operator for sufficiently large inputs at moderate levels of sparsity (around
50\%). For the smaller embedding dimension $D=384$, speedups are achieved even
at 10\% sparsity. At 90\% sparsity we achieve around 140\% peak floating
point 32 performance. Note that we, of course, calculate throughput in regard
to the dense number of floating point operations.
While saving memory is the main focus of this work,
faster execution is a welcome side effect.

\section{Activation Pruning}
\label{sec:pruning}

\begin{figure}[!t]
  \centering
  \includegraphics{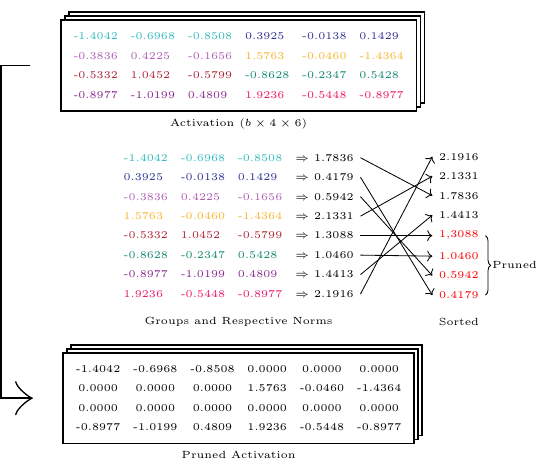}
  \caption{Example top-k block pruning step with block size \(3\) at 50\%
  sparsity. We calculate the \(l2\)-norm per block and select the lowest
  scoring blocks using top-k. Blocks are only compared within a single
  activation, not among others in the mini-batch.}\label{fig:pruning}
\end{figure}

In this work we use a simple magnitude-based top-k pruning approach to
sparsify activations. To force blocks of activations to zero, their vector or
matrix norms are compared and ranked. We then proceed to zero out the $k$
smallest blocks, where $k$ is determined by multiplying the total number of
blocks $N$ by the sparsity parameter $s$. The $l2$-norm is chosen to weigh
larger individual values stronger and retain a larger total norm of the
activation tensor, see Fig.~\ref{fig:pruning}. 

Blocks are only compared locally, not among other
activations in the mini-batch, because there is no direct relationship. For
example, one could imagine a darker image being pruned completely otherwise.
This makes the comparison computationally inexpensive. Additionally, this also
causes all activations in a mini-batch to have the same amount of zero and
non-zero blocks, reducing the BSR encoding overhead.

At this time, pruning is performed equally among all layers. For instance,
setting sparsity to 50\% will force all linear layers in the network to prune
50\% of their activations. As other work suggests that different layers have
different sensitivity to perturbations~\cite{Borras2022,Klein2022}, future work
will analyze each layer's sensitivity to pruning during training.

\section{Experiments}\label{sec:experiments}

We evaluate the feasibility of block-sparse training on the ResMLP-S12
architecture in the context of image classification. Due to time and compute
resource constrains, a full parameter grid search over the ImageNet dataset is
not viable. We therefore gauge the network's sensitivity to block-sparsity
using the smaller CIFAR10 and CIFAR100 datasets first. Based on the results, we
selectively test our method on ImageNet.

\subsection{Grid Search CIFAR10/100}\label{sec:experiments_c10}

\begin{figure*}[!t]
  \centering
  \subfloat[]{\includegraphics{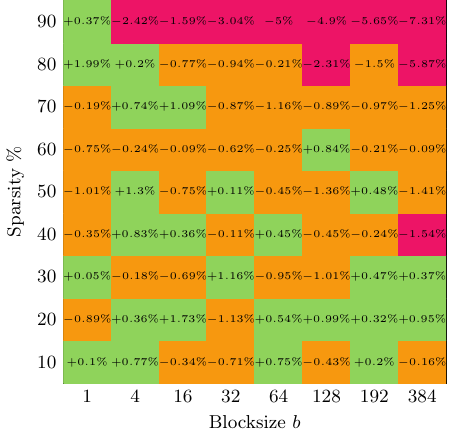}}
  \hspace{0.1\textwidth}
  \subfloat[]{\includegraphics{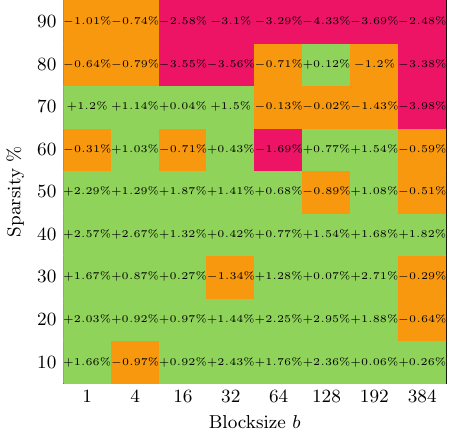}}
  \caption{Heatmap showing the top-1 accuracy relative to dense training for
  ResMLP-S12 on CIFAR10 (a) and CIFAR100 (b). Values above baseline accuracy are shown in
  green. Based on the observed standard deviation for dense training we define
  an arbitrary level of 1.5\% below baseline as acceptable. Acceptable values
  are shown in orange, while accuracies lower than 1.5\% below baseline are
  shown in red.}\label{fig:heatmap}
\end{figure*}

Fig.~\ref{fig:heatmap} shows the top-1 accuracy achieved relative to a densely
trained network for various configurations with regard to amount of sparsity
and block size. Besides these two parameters, all experiments share the same
hyperparameters\footnote{Hyperparameters are adapted from the original paper~\cite{touvronResMLPFeedforwardNetworks2021}}.

As to be expected, higher levels of sparsity lead to a decrease in accuracy.
This effect is stronger at larger block sizes.
Intuitively, pruning larger blocks will lead to "important" values being
pruned more often. Interestingly, low levels of sparsity can actually improve
accuracy, possibly due to pruning noise acting as a regularizer:
as smaller components will no longer contribute, parameter
updates become smaller in general.

Similarly, the effects of the block size become more pronounced
at higher levels of sparsity. Therefore, smaller block sizes are preferred.
For low levels of sparsity, there is no discernible trend.
This preference towards smaller block sizes is in direct opposition to our goal
of compression. However, viable configurations for $b \geq 16$ can be found up
to $70\%$ to $80\%$ of sparsity.

There is some statistical variation in the results. Some configurations perform
inexplicably well, like $(s=80\%, b=1)$ for CIFAR10, or badly, like $(s=30\%,
b=32)$ for CIFAR100. For reference, running the dense baseline configuration
with different weight initialization seeds shows a standard deviation of circa
$\pm1\%$ in top-1 accuracy. Future work should therefore encompass multiple runs
per configuration for statistical analysis.

\begin{figure}[!t]
  \centering
  \includegraphics{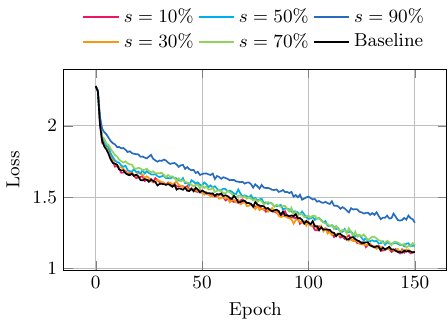}
  \caption{Comparing loss curves at increasing levels of sparsity for ResMLP-S12/CIFAR10.
  $b=384$ was chosen, as it produces the most extreme split. Because of the
  dense forward computation, the loss does not become more noisy, only the rate
  of convergence of the loss curve changes.}\label{fig:loss_384}
\end{figure}

Observing the loss curves for different configurations shows the training to
remain stable in terms of spread of the loss curve, see
Fig.~\ref{fig:loss_384}. $b=384$ is chosen as it produces the largest variation
in results. Due to the dense forward computation an accurate loss value is
obtained. However, because gradient calculation is lossy, parameter updates
become less accurate, affecting the rate of convergence. We can observe the
loss curves diverging within the first 10 epochs and running almost in parallel
afterward.

\subsection{Early Scaling Experiments}\label{sec:experiments_imagenet}

\begin{figure}[!t]
  \centering
  \includegraphics{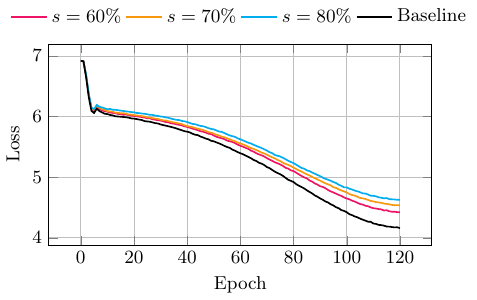}
  \caption{ResMLP-S12/ImageNet loss for $b=16$.
  A behavior similar to Fig.~\ref{fig:loss_384} can be observed. Again, only
  the rate of convergence is impacted by sparsity. The effect is more severe,
  however.}\label{fig:loss_imagenet}
\end{figure}

Using the results shown in Fig.~\ref{fig:heatmap} as a heuristic, we evaluate
block sparse training performance on the ImageNet dataset next. In
Fig.~\ref{fig:loss_imagenet} we inspect the loss curves, yielding results
similar to CIFAR10 and CIFAR100, with a more drastic spread, however.
Translating loss to accuracy, $s=60\%$ causes a reduction by $-5\%$ in
accuracy, $s=70\%$ causes $-7\%$, and $s=80\%$ leads to $-9\%$ top-1 validation
accuracy. However, based on the trends of the loss curves we expect slightly
longer training with an adapted learning rate schedule to improve accuracy to
near baseline. Further parameter studies are needed of course to conclude this
expectation.

To quantify the actual memory saved we additionally log memory allocations via
a custom allocator. Table~\ref{tab:memory_saved} shows the absolute and
relative memory consumed by activations for ImageNet training at mini-batch
size 32 for selected potential configurations.

\begin{table}[!t]
    \centering
    \caption{Memory Saved by Pruning (ResMLP-S12/ImageNet, Mini-Batch Size of 32)}
    \label{tab:memory_saved}
    \begin{tabular}{l cccc}
      \toprule
      Sparsity  & Block Size & Activations       & $\delta$          & $\delta$ \\
      $[\%]$    & --         & [\si{\mebi\byte}] & [\si{\mebi\byte}] & $[\%]$ \\
      \midrule
      Dense & n.a. & 1599.2 & --     & --   \\
      60    & 16   & 1234.0 & -365.2 & -22.8 \\
      60    & 32   & 1215.0 & -384.2 & -24.0 \\
      60    & 64   & 1205.3 & -393.9 & -24.6 \\
      70    & 16   & 1161.5 & -437.7 & -27.4 \\
      70    & 32   & 1142.9 & -456.3 & -28.5 \\
      70    & 64   & 1137.2 & -462.0 & -29.0 \\
      80    & 16   & 1082.8 & -516.4 & -32.3 \\
      80    & 32   & 1075.1 & -524.1 & -32.8 \\
      80    & 64   & 1071.0 & -528.2 & -33.0 \\
      \bottomrule
    \end{tabular}
\end{table}

\section{Discussion}

\begin{figure}[!t]
  \centering
  \includegraphics{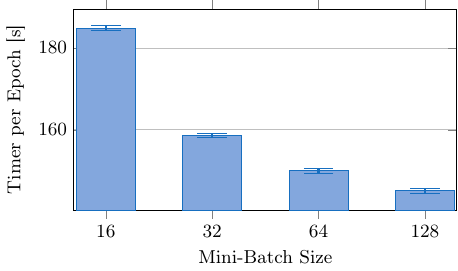}
	\caption{ResMLP-S12/CIFAR10 time per epoch versus mini-batch size. At lower
	mini-batch sizes the GPU is not utilized optimally, increasing the
	latency.}\label{fig:tbatch}
\end{figure}

\textbf{Mini-batch size}: A seemingly straight forward alternative to pruning is reducing the mini-batch
size. While directly reducing the memory footprint, doing so comes with
performance and, moreover, prediction accuracy implications.
Fig.~\ref{fig:tbatch} shows the latency of training ResMLP-S12/CIFAR10 for one
epoch at various mini-batch sizes. For small inputs there is not enough work
for the throughput oriented GPU to hide latency. Consequentially, training
becomes slower. At larger mini-batch sizes the opposite holds and the training
time decreases.

\begin{figure}[!t]
  \centering
  \includegraphics{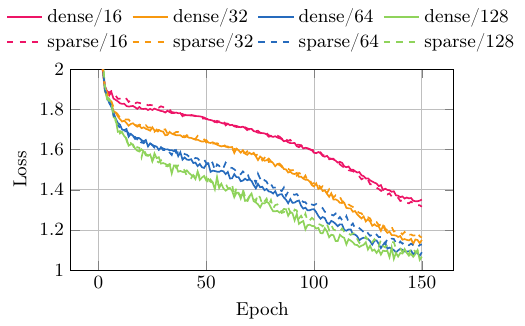}
	\caption{ResMLP-S12/CIFAR10 loss at various mini-batch sizes for dense and
	sparse training at $s=50\%$. The sparse loss follows the corresponding dense
	curve very closely performing better than the next smaller mini-batch size.}
  \label{fig:loss_batchsize}
\end{figure}

Comparing the training behavior of sparsely and densely trained network at
different mini-batch sizes, shows the sparse training to be much closer to
dense training at the same mini-batch size. This is shown in
Fig.~\ref{fig:loss_batchsize}, where we compare a densely trained ResMLP-S12 on
CIFAR10 to training with 50\% sparsity. This comparison is not entirely fair,
as 50\% sparsity does not translate to 50\% less total memory, but based on the
observations from Fig.~\ref{fig:heatmap} there is potential to move to higher
levels of sparsity without sacrificing accuracy. In this regard,
Table~\ref{tab:acc_batchsize} compares the achieved top-1 validation accuracies
for different mini-batch sizes. We can see that, on top of training
significantly slower, the network loses 7\% accuracy going from mini-batch size
32 to 16. Meanwhile, the accuracy for the sparse network at mini-batch size 32
does not differ from the dense network.
This could be a valuable alternative in resource-constrained systems, where a
viable mini-batch could otherwise not be used due to memory constraints.

\begin{table}[!t]
  \centering
  \caption{Comparing ResMLP-S12/CIFAR10 Accuracies at Different Mini-Batch
  Sizes to Sparse Training}\label{tab:acc_batchsize}
  \begin{tabular}{lcccc}
    \toprule
    Mini-Batch Size & 16 & 32 & 64 & 128 \\
    Model & \multicolumn{4}{c}{Top-1 Accuracy [\%]} \\
    \midrule
    dense & 82 & 89 & 91 & 90 \\
    50\% sparse & 82 & 89 & 90 & 90 \\
    \bottomrule
  \end{tabular}
\end{table}

\textbf{Overparameterization}: Although reasons for this are not yet fully
understood, overparameterization of networks, in combination with
regularization, seems to lead to better
generalization~\cite{zhangUnderstandingDeppLearning2021,kaplanScalingLawsNeural2020,liTrainBigThen2020}.
Our method can be used in this context to increase the model's complexity to
leverage this observation, counteracting the normally associated increase in
memory consumption to a degree.

While block sparse training applies to CIFAR10 and CIFAR100 without
repercussions in terms of accuracy and without any changes to the training
configuration, it seems that, in this configuration, further hyperparameter
tuning is required for ImageNet. It is however noteworthy, that at 15 million
parameters ResMLP-S12 is a relatively small architecture in the context of the
ImageNet dataset. Overparameterization is therefore expected to be low, thus
a larger variant is expected to show better results.

\textbf{Operator implementation}: We are also currently working on porting the
operator implementation from CUDA to
Triton\cite{tilletTritonIntermediateLanguage2019} for better portability and
extensibility. A block-sparse version of the affine scaling layer found in
ResMLP is now also available, potentially allowing for further memory savings,
and a convolution operator is planned. A block-sparse convolution operator
allows for interesting fundamental comparisons between vision MLPs and CNNs, as
the former emulate some of the properties found in the latter. The
cross-channel layer mimics the cross-channel property found in plain
convolutions. Similarly, the cross-patch layer can be seen as a 1D convolution,
mixing information among patches. Effects of differences, like weight sharing,
on robustness to block sparsity remain to be observed. CNNs also make for an
interesting candidate, as hardware friendly block sizes are relatively small.
The most typical kernel sizes in computer vision are $3 \times 3$ or $5 \times
5$. Due to the multitude of parameters, implementing an efficient block sparse
convolutional operator is challenging, however. Nvidia's cuDNN library, for
example, employs different algorithms depending on the
parameters\footnote{\url{https://docs.nvidia.com/deeplearning/performance/dl-performance-convolutional/index.html},
Section 1} via autotuning. Triton might also help in this regard, reducing
development time and increasing code readability.

\textbf{Stochasticity}: Related work~\cite{alistarhQSGD2017} on gradient
compression shows stochastic components to compression to be beneficial.
Therefore, future work will also encompass random components in the pruning
process. One could imagine randomly swapping blocks near the top-k threshold,
resulting in some blocks above the threshold being pruned anyway, and
vice-versa below threshold.
 
\section{Summary}

We have shown block sparse activation pruning based on a simple top-k criterion
to be a viable method for compressing the memory footprint of neural
architectures at the example of ResMLP. For CIFAR10/100 we can save around 32\%
of the memory consumed by activations, which translates to 28\% of the total
memory footprint, without affecting accuracy significantly. We have
additionally demonstrated the scalability of our block-sparse linear operator,
maintaining decent latency and throughput for as low as 10\% sparsity for
adequate block sizes.

Future work will encompass the directions mentioned in the discussion, with an
emphasis on larger neural architectures, larger data sets, as well as advanced
pruning methods including stochasticity, second-order pruning methods, and more
sophisticated pruning schedules.

\section*{Acknowledgements}
This work is part of the "Model-Based AI" project, which is funded by the Carl Zeiss Foundation.

\bibliographystyle{IEEEtran}
\bibliography{main}

\clearpage

\end{document}